\documentclass[journal, letterpaper]{IEEEtran}

\usepackage{subfig}   
\usepackage{flushend}

\ifCLASSINFOpdf
  \usepackage[pdftex]{graphicx}    
\else

\fi

\usepackage{bm}
\usepackage{amsmath}
\usepackage{xcolor}
\begin{document}

\title{Gradient Step Plug-and-Play Model\\ for Dental Cone-Beam CT Reconstruction}

\author{
        Idris Tatachak,
        Luis Kabongo,
        Nicolas Papadakis,
        Xavier Ripoche,
        Simon Rit
\thanks{TATACHAK Idris and RIT Simon are with INSA‐Lyon, Universite Claude Bernard Lyon 1, CNRS, Inserm, CREATIS UMR 5220, U1294, F‐69373, Lyon, France. (email: simon.rit@creatis.insa-lyon.fr)}
\thanks{PAPADAKIS Nicolas is with Univ. Bordeaux, CNRS, Inria, Bordeaux INP, IMB, UMR 5251, F-33400 Talence, France. (email: nicolas.papadakis@math.u-bordeaux.fr)}
\thanks{RIPOCHE Xavier, KABONGO Luis, and TATACHAK Idris are with ACTEON Group, France.}
\thanks{Corresponding email: idris.tatachak@creatis.insa-lyon.fr}
}

\pagenumbering{gobble}

\maketitle

\begin{abstract}
The goal of this work is to reduce the effect of photon noise in dental cone-beam CT reconstruction. We consider an inverse problem formulation and develop a data-based prior. To this end, we simulate fan-beam acquisitions and add photon noise to the projection data. The prior is obtained by training a gradient-step denoiser using reconstructed simulated acquisitions. The trained model is integrated into a plug-and-play gradient-step algorithm to reconstruct images from simulated projections. Experiments on synthetic data demonstrate the denoising capabilities of the trained model, while qualitative evaluations on real images showcase the algorithm’s performance and generalization ability.

\end{abstract}


\IEEEpeerreviewmaketitle

\section{Introduction}

\IEEEPARstart{I}{n} computed tomography (CT), regularization plays a major role in improving the image quality of reconstructions from CT acquisitions. The reconstruction problem can be formulated as:

\begin{equation}\label{eq:problem}
    x^* \in \arg\,\min_x f(x) + \lambda g(x),
\end{equation}
with $f(x)$ the data fidelity term, and $g(x)$ the regularization term weighted by the regularization parameter $\lambda\geq 0$. In the case of iterative CT reconstruction, the data fidelity term is usually taken as a quadratic loss as follows:

\begin{equation}
f(x) = \frac{1}{2}||Ax-p||_2^2.
\end{equation}

Many techniques and regularization functions have been proposed to treat degraded data (due to under-sampling or limited angle scan) and correct artifacts (e.g. metal or photon starvation artifacts) present in CT data. These regularization techniques act as priors in our problem, injecting prior information about the object we aim to reconstruct. Recently,  significant contributions have been made in the direction of data-based priors used in the context of model-based iterative reconstruction (MBIR) algorithms. These approaches leverage a prior $g(x)$ that is either explicitly defined (e.g., total variation) or implicitly given by an available model like an image denoiser. With the development of deep learning, the number and effectiveness of implicit models have increased drastically in the past years.

A notable class of algorithms used in MBIR is Plug-and-Play (PnP) algorithms~\cite{venkatakrishnan2013plug,kamilov2023plug}. PnP algorithms use off-the-shelf image denoisers to incorporate an implicit prior in an iterative reconstruction algorithm. First proposed using classical denoisers such as BM3D~\cite{BM3D}, PnP methods have increasingly adopted deep denoisers \cite{zhang2017beyond,zhang2021plug}. 

The performance and convergence behavior of iterative PnP algorithms are inherently tied to the properties of the denoiser $D$ being employed.
A main question in the development of PnP models thus concerns the  properties required by the denoiser $D$ to ensure the convergence of the iterative process providing the reconstruction. Many constraints have been proposed such as 1-Lipschitz constraint of the denoiser~\cite{ryu2019plug}, which limits the network expressiveness, or firmly non-expansive constraint \cite{pesquet2021learning} that corresponds to a 1-Lipschitz operator  $(2D - \mathrm{Id})$, in order to ensure the convergence of the iterates. This last constraint leads to a more expressive network, but it is not associated to an explicit regularization function $g$.

In this work, we leverage on  
the \emph{Gradient Step Denoiser}~\cite{hurault2021} to build an iterative  CT reconstruction algorithm.
A key advantage of this denoiser is that it does not require an explicit penalization of the Lipschitz constant of its Jacobian, which can be challenging in practice~\cite{pesquet2021learning}. 
In this work, we first train a deep gradient step denoiser using simulated fanbeam acquisitions and demonstrate its performance on real dental cone-beam CT (CBCT) acquisitions. Section \ref{Methods} describes the simulated database generation process, the model training framework, and the reconstruction algorithm used. Section \ref{Results} presents reconstruction results obtained with our framework on real dental CT data. We finally conclude in section \ref{Conclusion}, pointing to potential improvements of the presented method.

\section{Materials and Methods} \label{Methods}
In this section, we  first present the Gradient Step Model used for regularization and then described the proposed approach, which  consists of three sequential steps: database generation, model training for the general denoising task (independent of reconstruction), and image reconstruction using the trained model.

\subsection{Gradient Step Denoiser}

The Gradient Step Denoiser~\cite{hurault2021} is designed to minimize an explicit regularization term $g$ defined as

\begin{equation}
g(x) = \frac12||x-N_\sigma(x)||^2,
\end{equation}
where $N_\sigma(x)$ is a differentiable image-to-image neural network depending on a noise level parameter $\sigma$, whose Jacobian is noted $J_{N_\sigma}$. The gradient of the regularization term $\nabla g$ is 

\begin{equation}
\nabla g(x) = x-N_\sigma(x) -  J_{N_\sigma}(x)^T(x-N_\sigma(x))
\end{equation}
and the final gradient step denoiser is defined as
\begin{equation} \label{eq:def_gradientstep}
    D_\sigma = \mathrm{Id} - \nabla g=N_\sigma(x)  + J_{N_\sigma}(x)^T(x-N_\sigma(x)).
\end{equation}
This denoiser therefore modifies the original architecture $N_\sigma$ to ensure it represents the gradient of an explicit function. This is a crucial property guaranteeing convergence when integrated into an iterative reconstruction process~\cite{hurault2021}.

\subsection{Data Generation}

PnP models are usually trained on natural images, which has the advantage of providing a large amount of training data. Here, we train the model on images close to the distribution of real dental CT images to learn a prior that is better suited to our problem. 
To this end, we performed simulations of polychromatic fan-beam acquisitions of 2D slices of the XCAT phantom \cite{XCAT} using RTK \cite{RTK_2}. Noise was then added to the simulated projections following a compound Poisson distribution, where the rate parameter was estimated from air acquisitions of real CBCT projections.

As a result, our dataset is constructed to provide paired examples required for training the model:
\begin{itemize}
\item \textbf{Input:} a reconstructed image obtained from a noisy acquisition.
\item \textbf{Ground truth:} the corresponding reconstructed image obtained from a noise-free acquisition.
\end{itemize}

\subsubsection{Phantom Preparation} Simulations were performed using the XCAT phantom \cite{XCAT}, which provides a full-body material map. In total, 1490 slices of the XCAT phantom were extracted, each with a thickness of 1.1 mm, starting from the bottom of the phantom. Each slice has an in-plane pixel spacing of 0.14 mm, resulting in images of size $2092 \times 2092$ pixels.

To augment the dataset, patches of size $888 \times 888$ pixels were extracted from these slices. The patches were selected such that the field-of-view (FOV) contained at least $20\%$ non-air material and at least $10\%$ air, in order to  avoid the interior problem (where projections are truncated on both sides, leading the solution to be non-unique \cite{interior_pb}). Patches not satisfying these criteria were discarded, resulting in a total of 5,715 samples.

\subsubsection{CT simulation} Acquisitions were simulated using the central slice geometry of a dental CBCT scanner. Line integrals were computed using the RTK CudaRayCast operator on the different materials $m$ (e.g. water, muscle, cartilage, skull bone, ...), and a given machine spectrum. The measured intensity is computed as:

\begin{equation}
I = \sum_E S(E)\,\exp\!\left(-\sum_m \int_L \mu_m(x, E)\, \mathrm{d}x\right)
\end{equation}
where $S(E)$ denotes the effective spectrum (emitted spectrum multiplied by the detector sensitivity) at energy level $E$, sampled from 0 to 90 keV in 0.5 keV intervals, for a tube voltage of 90 kV and 12.5~mAs. The index $m$ denotes a material from the set of materials defined in the XCAT phantom.

\subsubsection{Noise model} The noise is restricted to photon noise without electronic noise as it is dominant at the selected dose level. The photon noise in the simulated projections is modeled using a compound Poisson distribution resulting from the sum of Poisson realizations at each energy level,

\begin{equation}
N \sim  \sum_E  \mathcal{P}\left(S(E) \exp\!\left(-\sum_m \int_L \mu_m(x, E)\, \mathrm{d}x\right)\right).
\end{equation}

\subsubsection{Reconstruction and ground-truth}

From the aforementioned simulated acquisitions, fan-beam CT images were reconstructed using the same geometry with $592 \times 592$ pixels with 0.15~mm spacing. 
Truncation of the projections resulting from the patient being larger than the FOV, common in dental CBCT, was handled by correcting the projections before reconstruction. The truncation correction was performed according to the approach based on the subtraction of the out-of-FOV forward projection of a coarse grid CT image, as detailed in \cite{Dang2016,cropped}, with a coarse grid of $592 \times 592$ pixels of 0.4~mm.

\begin{figure}[!t]
\centering
\includegraphics[width=3.5in]{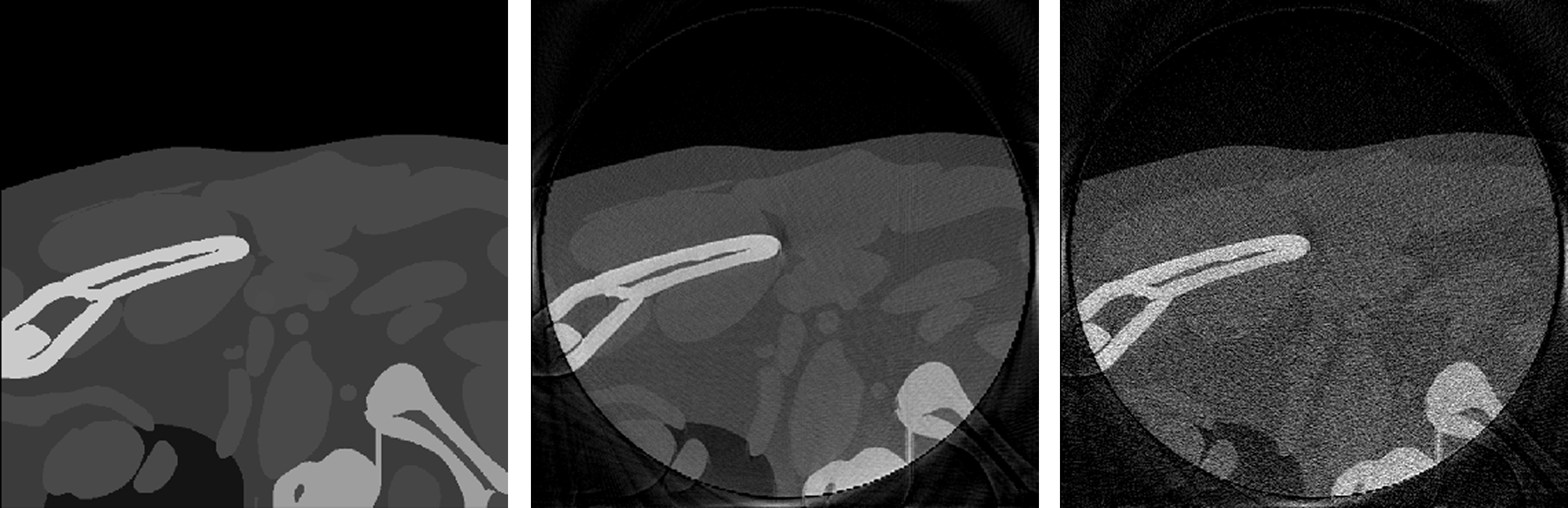}
\caption{Example slices of the XCAT phantom, displayed with the same window (0, 0.15) $mm^{-1}$. Left: Effective attenuation map of the region to be reconstructed, computed as the linear attenuation coefficient at the average spectrum energy. Center: Reconstructed image from simulated acquisitions with no simulated noise. Right: Idem with noise.  }
\label{fig_sim}
\end{figure}

For the ground truth, we used the reconstructed images from acquisitions without simulated noise (see Fig. \ref{fig_sim}). This choice was made instead of using the effective attenuation map in order to prevent the model from learning the correction of beam hardening and truncation artifacts. 

\subsubsection{Real acquisitions} Real dental CBCT acquisitions were used to qualitatively validate the reconstruction algorithm. No pre-processing or post-processing was applied to the data. In particular, calibration, dead-pixel correction, and patient motion correction were not applied to the examples presented here. Because the study focuses on 2D denoising, slices were extracted from the volumes to get 2D data. 

\begin{figure*}[!htbp]
\centering
\includegraphics[width=7in]{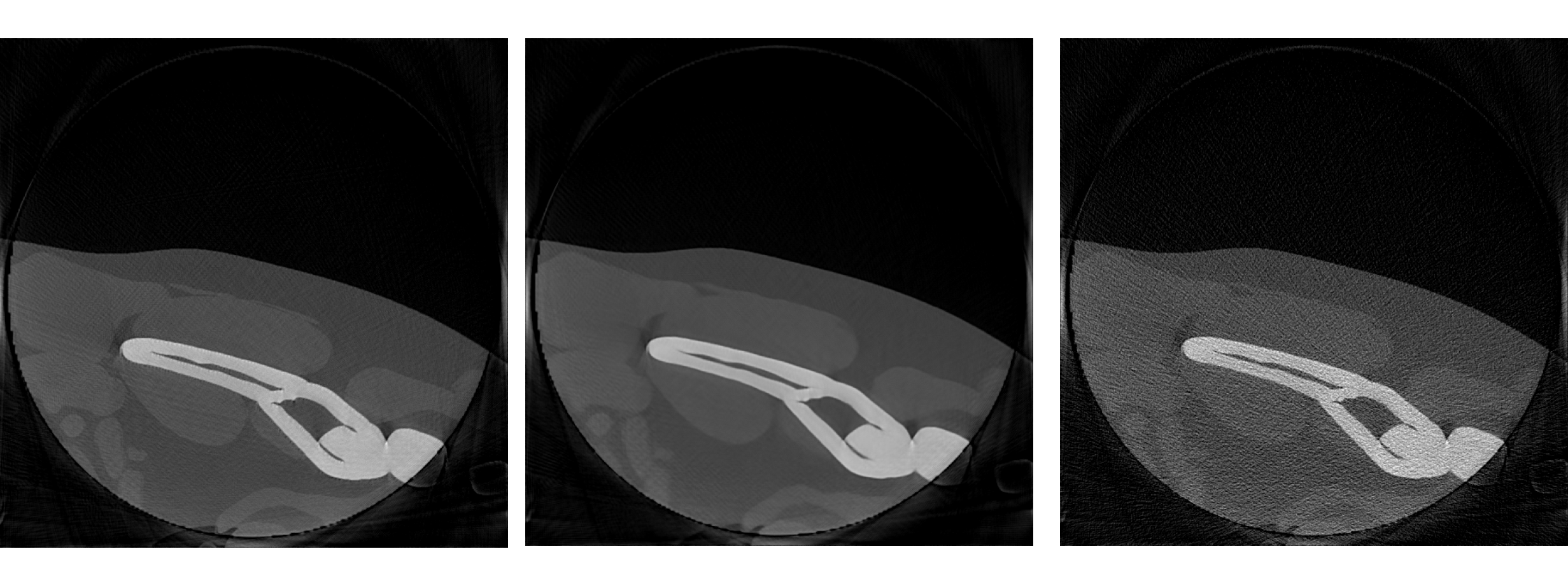}
\caption{
Example of application of the denoiser on a denoising task for a simulated test-set image at the jaw level, in post-processing. 
Left: Reference image. 
Middle: Image denoised by the trained model (PSNR = 34.77~dB). 
Right: Noisy image (PSNR = 23.49~dB).
}
\label{fig_post}
\end{figure*}

\subsection{Network and training}

We trained a denoiser based on the DRUNet architecture \cite{zhang2021plug}, made of four blocks, with one input channel and one output channel, for an image denoising task.

The dataset of 5715 images was divided into 4317 training samples ($\sim$ 76\%), 600 validation samples ($\sim$ 10\%), and 743 test samples ($\sim$14\%), the jaw region of the XCAT phantom being reserved exclusively for the test set, as it is our region of interest.

To initialize the model weights, we used the pretrained weights from~\cite{hurault2021}. Additionally, we replaced the DRUNet's activation layer with a softplus function instead of ELU, to ensure continuous differentiability with respect to the input \cite{pmlr-v162-hurault22a}.

The model was fine-tuned for 63 epochs with a learning rate of $5\times 10^{-5}$ and a batch size of 8. The loss consists of the L1 distance between the network output and the ground-truth image:

\begin{equation}
\mathcal{L} (x, \hat{x}) = |x-\hat{x}|.
\end{equation}

\subsection{Gradient Step Plug-and-Play (GS-PnP) reconstruction algorithm}
To reconstruct images from the simulated acquisitions, we solved problem~\eqref{eq:problem} with an iterative reconstruction algorithm, the steepest gradient descent:
\begin{equation}
    x_{k+1} = x_k - \tau \big(\nabla f(x_k) + \lambda \nabla g(x_k)\big),
\end{equation}
where $\tau$ denotes the step size, and $\lambda$ the regularization parameter. The regularization gradient $\nabla g$ is computed at each iteration following the denoiser defined in relation~\eqref{eq:def_gradientstep}:

\begin{equation}
    \nabla g(x_k) = x_k - D_\sigma(x_k).
\end{equation}

The algorithm was initialized with $x_0=\bm{0}$ and ran for 1500 iterations. The algorithm was implemented using DeepInverse~\cite{tachella2025deepinverse}. Several values of the regularization parameter \(\lambda\) were tested to evaluate their effect on reconstruction quality.

\section{Results} \label{Results}

We first report the results of model training on the denoising task in Subsection \ref{TestDenoiser}. Then, we present in Section \ref{TestPnP} the reconstruction results obtained using the trained denoiser.

\subsection{Model Testing} \label{TestDenoiser}

The trained denoiser was evaluated on a test set of 798 images obtained from the simulated acquisition of the XCAT phantom. The model was directly applied as a post-processing step to the reconstructed noisy images to assess its performance in this task. 
The model achieved a mean PSNR of 34.7~dB on the test set when comparing the denoised output with the non-noisy reconstruction over the entire image, whereas the noisy input achieved a PSNR of 22.95~dB.  

Figure \ref{fig_post} presents a qualitative illustration in which the denoised image (Fig. \ref{fig_post}, middle) appears noticeably cleaner while preserving the same contrast as in the clean reconstruction (Fig. \ref{fig_post}, left). However, some fine structures visible in the ground truth, such as the small circular feature to the left of the jawbone, were obscured by noise in the input image (Fig. \ref{fig_post}, right) and were not fully recovered by the model.

\begin{figure*}[!htbp]
\centering
\includegraphics[width=7.0in]{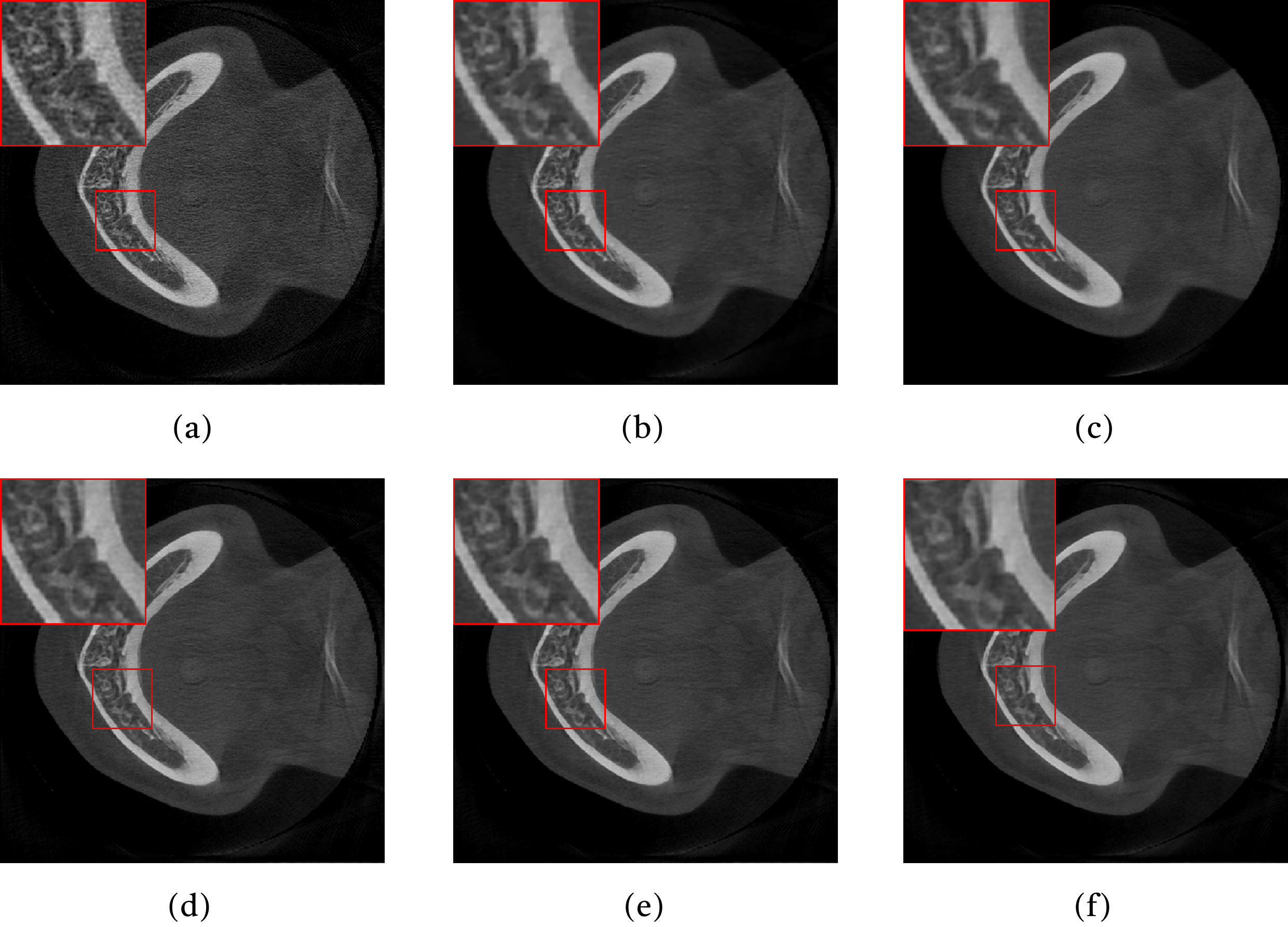} 
\caption{Example of application of the Gradient-Step denoiser as a prior in a reconstruction task from a real acquisition, using 1500 iterations, compared to TV and a Denoiser from \cite{zhang2021plug}. 
$\textbf{(a)}$ Reconstruction without prior using gradient descent ($\lambda = 0$). 
$\textbf{(b)}$ Reconstruction using a TV prior with PGD; over 20 iterations, with a regularization parameter of $\lambda = 0.05$. 
$\textbf{(c)}$ Reconstruction using a DRUNet prior with $\alpha$-PGD; the denoiser replacing the proximal operator used an averaging parameter $\alpha = 0.004$. 
$\textbf{(d)}$ Reconstruction using the trained model as a prior with $\lambda = 15$. 
$\textbf{(e)}$ Same as (d), but reconstructed with $\lambda = 20$.
$\textbf{(f)}$ Same as (d), but reconstructed with $\lambda = 30$.\vspace{-0.1cm}
}
\label{fig_real}
\end{figure*}

\subsection{Performance on PnP Reconstruction} \label{TestPnP}

A quantitative evaluation was performed by testing the GS-PnP algorithm on a CT reconstruction task. The evaluation used 258 simulated acquisitions from the test set in the jaw region of the XCAT phantom. The ground truth corresponded to reconstructions without simulated noise, and the regularization parameter was set to $\lambda = 30$. We obtained a mean PSNR of 32.1~dB compared to 22.2~dB without the application of the prior obtained with the conjugate gradient algorithm.

Qualitative results were obtained by applying the GS-PnP algorithm to reconstruct real acquisitions extracted from CBCT slices. These results were compared with reconstructions obtained using (i) a Total Variation (TV) prior, and (ii) the Gaussian denoiser from~\cite{zhang2021plug} plugged as a proximal operator in an $\alpha$-Proximal Gradient Descent ($\alpha$-PGD) scheme~\cite{hurault2023relaxed}. We used the Gaussian denoiser  provided by the DeepInverse library~\cite{tachella2025deepinverse}, which has been trained on natural images.

Tests were performed using different values for the hyperparameter $\lambda$ controlling the regularization strength. Figure~\ref{fig_real} illustrates that increasing the regularization parameter from $\lambda=15$ to $\lambda=30$ leads to a smoother result, though some fine details get attenuated by the model. We also observe than the results from our data-based model (Fig. \ref{fig_real}.d) is sharper than reconstruction from the TV prior (Fig. \ref{fig_real}.b) and the $\alpha$-PGD Gaussian denoiser (Fig. \ref{fig_real}.c).
Nevertheless, small low-attenuation regions located between high-attenuation areas  seem to suffer from reduced contrast, particularly for large values of $\lambda$.

\section{Conclusion} \label{Conclusion}

In this work, we trained a data-driven prior for CT reconstruction that effectively reduces photon noise in reconstructed images. The model is trained as a gradient-step denoiser, minimizing an explicit potential. Crucially, this potential is learned directly from the noise distribution encountered in CT images, rather than assuming Gaussian noise, making it better suited to the problem.

Challenges still persist in the reconstruction of some fine and clinically relevant structures. These limitations may be attributed to the use of a phantom with limited fine-detail features. More representative phantoms for training the denoiser could help improve reconstruction accuracy.

A major limitation of the currently presented approach is the slow convergence of the algorithm used, the steepest gradient descent being slow to converge. Future work will focus on implementing our approach using faster algorithms that are better suited to the reconstruction problem at hand.

\section*{Acknowledgment}

This work was supported by the French National Association for Technical Research (ANRT) and the France 2030 BPI project Cone Beam AI  and was performed within the framework of the SIRIC LYriCAN+ (INCa-DGOS-INSERM-ITMO cancer 18003) and the LABEX PRIMES (ANR-11-LABX-0063) of Université de Lyon, within the program ‘Investissements d’Avenir’ (ANR- 11-IDEX-0007) operated by the ANR. Simon Rit is a Royal Society Wolfson Visiting Fellow at the University College of London funded by the Wolfson Foundation and Royal Society.

\bibliographystyle{IEEEtran}

\begin{thebibliography}{}
\providecommand{\url}[1]{#1}
\csname url@samestyle\endcsname
\providecommand{\newblock}{\relax}
\providecommand{\bibinfo}[2]{#2}
\providecommand{\BIBentrySTDinterwordspacing}{\spaceskip=0pt\relax}
\providecommand{\BIBentryALTinterwordstretchfactor}{4}
\providecommand{\BIBentryALTinterwordspacing}{\spaceskip=\fontdimen2\font plus
\BIBentryALTinterwordstretchfactor\fontdimen3\font minus
  \fontdimen4\font\relax}
\providecommand{\BIBforeignlanguage}[2]{{%
\expandafter\ifx\csname l@#1\endcsname\relax
\typeout{** WARNING: IEEEtran.bst: No hyphenation pattern has been}%
\typeout{** loaded for the language `#1'. Using the pattern for}%
\typeout{** the default language instead.}%
\else
\language=\csname l@#1\endcsname
\fi
#2}}
\providecommand{\BIBdecl}{\relax}
\BIBdecl

\end{thebibliography}


\begin{thebibliography}{10}
\providecommand{\url}[1]{#1}
\csname url@samestyle\endcsname
\providecommand{\newblock}{\relax}
\providecommand{\bibinfo}[2]{#2}
\providecommand{\BIBentrySTDinterwordspacing}{\spaceskip=0pt\relax}
\providecommand{\BIBentryALTinterwordstretchfactor}{4}
\providecommand{\BIBentryALTinterwordspacing}{\spaceskip=\fontdimen2\font plus
\BIBentryALTinterwordstretchfactor\fontdimen3\font minus \fontdimen4\font\relax}
\providecommand{\BIBforeignlanguage}[2]{{%
\expandafter\ifx\csname l@#1\endcsname\relax
\typeout{** WARNING: IEEEtran.bst: No hyphenation pattern has been}%
\typeout{** loaded for the language `#1'. Using the pattern for}%
\typeout{** the default language instead.}%
\else
\language=\csname l@#1\endcsname
\fi
#2}}
\providecommand{\BIBdecl}{\relax}
\BIBdecl

\bibitem{venkatakrishnan2013plug}
S.~V. Venkatakrishnan, C.~A. Bouman, and B.~Wohlberg, ``Plug-and-play priors for model based reconstruction,'' in \emph{2013 IEEE global conference on signal and information processing}.\hskip 1em plus 0.5em minus 0.4em\relax IEEE, 2013, pp. 945--948.

\bibitem{kamilov2023plug}
U.~S. Kamilov, C.~A. Bouman, G.~T. Buzzard, and B.~Wohlberg, ``Plug-and-play methods for integrating physical and learned models in computational imaging: Theory, algorithms, and applications,'' \emph{IEEE Signal Processing Magazine}, vol.~40, no.~1, pp. 85--97, 2023.

\bibitem{BM3D}
K.~Dabov, A.~Foi, V.~Katkovnik, and K.~Egiazarian, ``Image denoising by sparse {3-D} transform-domain collaborative filtering,'' \emph{IEEE Transactions on Image Processing}, vol.~16, no.~8, pp. 2080--2095, 2007.

\bibitem{zhang2017beyond}
K.~Zhang, W.~Zuo, Y.~Chen, D.~Meng, and L.~Zhang, ``Beyond a {Gaussian} denoiser: Residual learning of deep {CNN} for image denoising,'' \emph{IEEE Trans. on Image Processing}, vol.~26, no.~7, pp. 3142--3155, 2017.

\bibitem{zhang2021plug}
K.~Zhang, Y.~Li, W.~Zuo, L.~Zhang, L.~Van~Gool, and R.~Timofte, ``Plug-and-play image restoration with deep denoiser prior,'' \emph{IEEE Trans. on Pattern Analysis and Mach. Intel.}, vol.~44, no.~10, pp. 6360--6376, 2021.

\bibitem{ryu2019plug}
E.~Ryu, J.~Liu, S.~Wang, X.~Chen, Z.~Wang, and W.~Yin, ``Plug-and-play methods provably converge with properly trained denoisers,'' in \emph{International Conference on Machine Learning}, 2019, pp. 5546--5557.

\bibitem{pesquet2021learning}
J.-C. Pesquet, A.~Repetti, M.~Terris, and Y.~Wiaux, ``Learning maximally monotone operators for image recovery,'' \emph{SIAM Journal on Imaging Sciences}, vol.~14, no.~3, pp. 1206--1237, 2021.

\bibitem{hurault2021}
S.~Hurault, A.~Leclaire, and N.~Papadakis, ``Gradient step denoiser for convergent plug-and-play,'' \emph{CoRR}, vol. abs/2110.03220, 2021.

\bibitem{XCAT}
W.~P. Segars, G.~Sturgeon, S.~Mendonca, J.~Grimes, and B.~M.~W. Tsui, ``{4D XCAT} phantom for multimodality imaging research,'' \emph{Medical Physics}, vol.~37, no.~9, pp. 4902--4915, 2010.

\bibitem{RTK_2}
S.~Rit, S.~Brousmiche, J.~Finet, G.~C. Sharp, and P.~Steininger, ``{Reconstruction Toolkit (RTK) v2, an Insight Toolkit (ITK) module for tomographic reconstruction},'' in \emph{{International Conference on the use of Computers in Radiation therapy (ICCR)}}, Lyon, France, 2024.

\bibitem{interior_pb}
\BIBentryALTinterwordspacing
K.~T. Smith, D.~C. Solmon, S.~L. Wagner, and C.~Hamaker, ``Mathematical aspects of divergent beam radiography,'' \emph{Proceedings of the National Academy of Sciences}, vol.~75, no.~5, pp. 2055--2058, 1978. [Online]. Available: \url{https://www.pnas.org/doi/abs/10.1073/pnas.75.5.2055}
\BIBentrySTDinterwordspacing

\bibitem{Dang2016}
H.~Dang, J.~W. Stayman, A.~Sisniega, and W.~Zbijewski~et al., ``\BIBforeignlanguage{en}{Multi-resolution statistical image reconstruction for mitigation of truncation effects: application to cone-beam {CT} of the head},'' \emph{\BIBforeignlanguage{en}{Phys Med Biol}}, vol.~62, no.~2, pp. 539--559, 2016.

\bibitem{cropped}
H.~S. Park and K.~Jeon, ``An iterative reconstruction method for dental cone-beam computed tomography with a truncated field of view,'' \emph{arXiv preprint arXiv:2508.07618}, 2025.

\bibitem{pmlr-v162-hurault22a}
S.~Hurault, A.~Leclaire, and N.~Papadakis, ``Proximal denoiser for convergent plug-and-play optimization with nonconvex regularization,'' in \emph{Int. Conference on Machine Learning}, vol. 162, 2022, pp. 9483--9505.

\bibitem{tachella2025deepinverse}
J.~Tachella, M.~Terris, S.~Hurault, and A.~Wang~et al., ``Deepinverse: A python package for solving imaging inverse problems with deep learning,'' \emph{J. of Open Source Software}, vol.~10, no. 115, p. 8923, 2025.

\bibitem{hurault2023relaxed}
S.~Hurault, A.~Chambolle, A.~Leclaire, and N.~Papadakis, ``A relaxed proximal gradient descent algorithm for convergent plug-and-play with proximal denoiser,'' in \emph{International Conference on Scale Space and Variational Methods in Computer Vision}.\hskip 1em plus 0.5em minus 0.4em\relax Springer, 2023, pp. 379--392.

\end{thebibliography}

\end{document}